\documentclass[a4paper,12pt]{article}
\usepackage{CJKutf8}
\usepackage[super,square]{natbib}
\usepackage{anysize}
\marginsize{2cm}{2cm}{1cm}{2cm}
\usepackage{fancyhdr}

\usepackage{soul}
\usepackage[dvipsnames]{xcolor}
\usepackage{hyperref}
\usepackage{graphicx}
\hypersetup{
    colorlinks=true,
    linkcolor=black,
    citecolor=CadetBlue,
    filecolor=CadetBlue,      
    urlcolor=CadetBlue,
}
\usepackage{lipsum}
\usepackage{multicol}
\usepackage{parskip}
\renewenvironment{abstract}
 {\par\noindent\textbf{\abstractname}\ \ignorespaces \\}
 {\par\noindent\medskip}

\begin{document}
\begin{CJK}{UTF8}{gbsn}
\pagestyle{fancy}
\thispagestyle{empty}
\fancyhead[R]{}
\fancyhead[L]{}
\renewcommand*{\thefootnote}{\fnsymbol{footnote}}
\begin{center}
\Large{\textbf{A Survey on GUI Agents with Foundation Models Enhanced by Reinforcement Learning}}
\vspace{0.4cm}
\normalsize
\\ Jiahao Li, Kaer Huang \\
\vspace{0.1cm}
\textit{Lenovo Research}\\
\vspace{0.1cm}
\textit{jiahaoli0301@gmail.com}
\medskip
\normalsize
\end{center}
{\color{gray}\hrule}
\vspace{0.4cm}
\begin{abstract}
Graphical User Interface (GUI) agents, driven by Multi-modal Large Language Models (MLLMs), have emerged as a promising paradigm for enabling intelligent interaction with digital systems. This paper provides a structured summary of recent advances in GUI agents, focusing on architectures enhanced by Reinforcement Learning (RL). We first formalize GUI agent tasks as Markov Decision Processes and discuss typical execution environments and evaluation metrics. We then review the modular architecture of (M)LLM-based GUI agents, covering Perception, Planning, and Acting modules, and trace their evolution through representative works. Furthermore, we categorize GUI agent training methodologies into Prompt-based, Supervised Fine-Tuning (SFT)-based, and RL-based approaches, highlighting the progression from simple prompt engineering to dynamic policy learning via RL. Our summary illustrates how recent innovations in multimodal perception, decision reasoning, and adaptive action generation have significantly improved the generalization and robustness of GUI agents in complex real-world environments. We conclude by identifying key challenges and future directions for building more capable and reliable GUI agents.
\end{abstract}
{\color{gray}\hrule}
\medskip
\begin{multicols}{2}
\tableofcontents
\section{Introduction}
With the continuous advancements of LLMs in the fields of Natural Language Processing (NLP) and Computer Vision (CV), LLMs have demonstrated impressive performance in various downstream tasks after being trained on large-scale corpora using the "next token prediction" paradigm\cite{LLaMA}\cite{GPT}\cite{qwen}. LLMs have progressively evolved from a simple conversational chatbot\cite{llama2openfoundation}\cite{survey1}\cite{Chatbots} to autonomous agents\cite{sparks}\cite{Wang_2024}\cite{chatdev} capable of planning\cite{cot}, tool use\cite{tools}, and memory management\cite{park2023}\cite{EI} to perform and complete complex tasks. Building on this foundation, a series of works have emerged that utilize LLMs to interact with digital systems\cite{MP-GUI}\cite{CogAgent}\cite{uihawk}\cite{UI-R1}\cite{UI-TARS}\cite{GUI-R1}\cite{Falcon-UI}\cite{GUICourse}\cite{UFO2}\cite{InfiGUIAgent}\cite{Mobile-Agent-v2}(e.g., smartphones or computers), where LLMs interact with digital devices through Graphical User Interfaces (GUIs) in the same way as humans.

This paradigm holds significant research and development potential, as GUIs are integral to nearly all digital devices, smartphones, tablets, desktops, and even televisions, used in daily human activities such as work, learning, and leisure. In the era of rapid AI development, using LLM-based agents to control digital devices to fulfill human user needs can significantly enhance the user experience, offering more convenient and efficient services, and demonstrating great value potential. However, previous studies based on traditional rule-based and RL methods have struggled with tasks resembling human interactions\cite{liu2018}, limiting their applicability.

Recent advances in LLMs and MLLMs have significantly bolstered their capabilities in semantic comprehension and cognitive reasoning. Agents built on (M)LLMs can effectively interpret and integrate textual and visual input, enabling them to devise detailed strategies to tackle complex tasks. These breakthroughs also provide opportunities to address previously mentioned challenges in handling complex tasks, making it possible for agents to autonomously complete user tasks in GUIs.

This paper aims to summarize recent learning experiences and provide an overview of the current novel and influential approaches for GUI agents. Specifically, it covers the foundational concepts of GUI agents, the classification of execution environments, architectural frameworks, and key methodological categories.
\end{multicols}
{\color{gray}\hrule}
\begin{center}
\section{Problem Formulation}
\bigskip
\end{center}
{\color{gray}\hrule}
\begin{multicols}{2}
A GUI agent refers to an agent that autonomously interacts with digital devices (such as smartphones or desktop computers) through GUIs. It makes decisions and generates plans based on user-defined task instructions and the current screen context, executing actions through interaction with clickable or typeable elements to achieve user goals.

The interaction process between GUI agents and their environments is commonly formalized as a Markov Decision Process (MDP), denoted as \(M = \{S, A, T, r,\gamma \}\) . In this formulation, S represents the state space, defined as the set of all possible screen captures of the digital device. A denotes the set of permissible actions, such as clicking, typing, and other forms of interaction on the screen. The transition function $T:S\!\times\!A\!\times\!S\!\rightarrow\![0, 1]$ defines the probability distribution over the next states given a current state and an action. The reward function $r:S\!\times\!A\!\rightarrow\!R$ assigns a scalar feedback value to each state-action pair. \(\gamma\) is the discount factor. The goal of the GUI agent is to learn a decision policy \(\pi\) that maximizes returns. We typically represent \(\pi(a|s)\!\in\![0, 1]\) as the probability of selecting action a in state s under the policy \(\pi\).
\end{multicols}
{\color{gray}\hrule}
\begin{center}
\section{Benchmarks}
\bigskip
\end{center}
{\color{gray}\hrule}
\begin{multicols}{2}
The execution environments of GUI agents usually include static environment datasets and dynamic interactive environments. In static settings, the environment remains constant, allowing GUI agents to work within predefined datasets without needing to account for environmental changes in action planning. In dynamic environments, the state may evolve after each action, necessitating that agents observe the updated state to determine subsequent actions. Dynamic environments are generally classified into simulated and real-world settings. Simulated environments are usually cleaner, free from disturbances like pop-up ads, and provide a standardized interactive setting for testing. In real environments, the agent must observe and interact with the digital device like humans. Although this setup more accurately reflects real-world scenarios, it also exposes agents to numerous unpredictable factors.

The success rate is the most common metric used to evaluate a GUI agent, reflecting its ability to complete tasks effectively. In addition, execution efficiency is another common evaluation metric, often measured by the number of steps, time, or cost required to complete a task. Notably, for dynamic interactive environments, due to uncertainty from environmental changes, the fixed evaluation metrics mentioned earlier may not be applicable. As a result, human experts are required for manual assessment, which compromises reproducibility and requires significant labor costs. Consequently, recent research has explored leveraging MLLMs for automating the evaluation process, aiming to reduce human involvement. However, due to the hallucination issues in LLMs, ensuring evaluation accuracy while reducing human effort remains a critical research challenge.
\end{multicols}

\begin{figure*}[thb]
    \centering
    \includegraphics[width=0.99\textwidth]{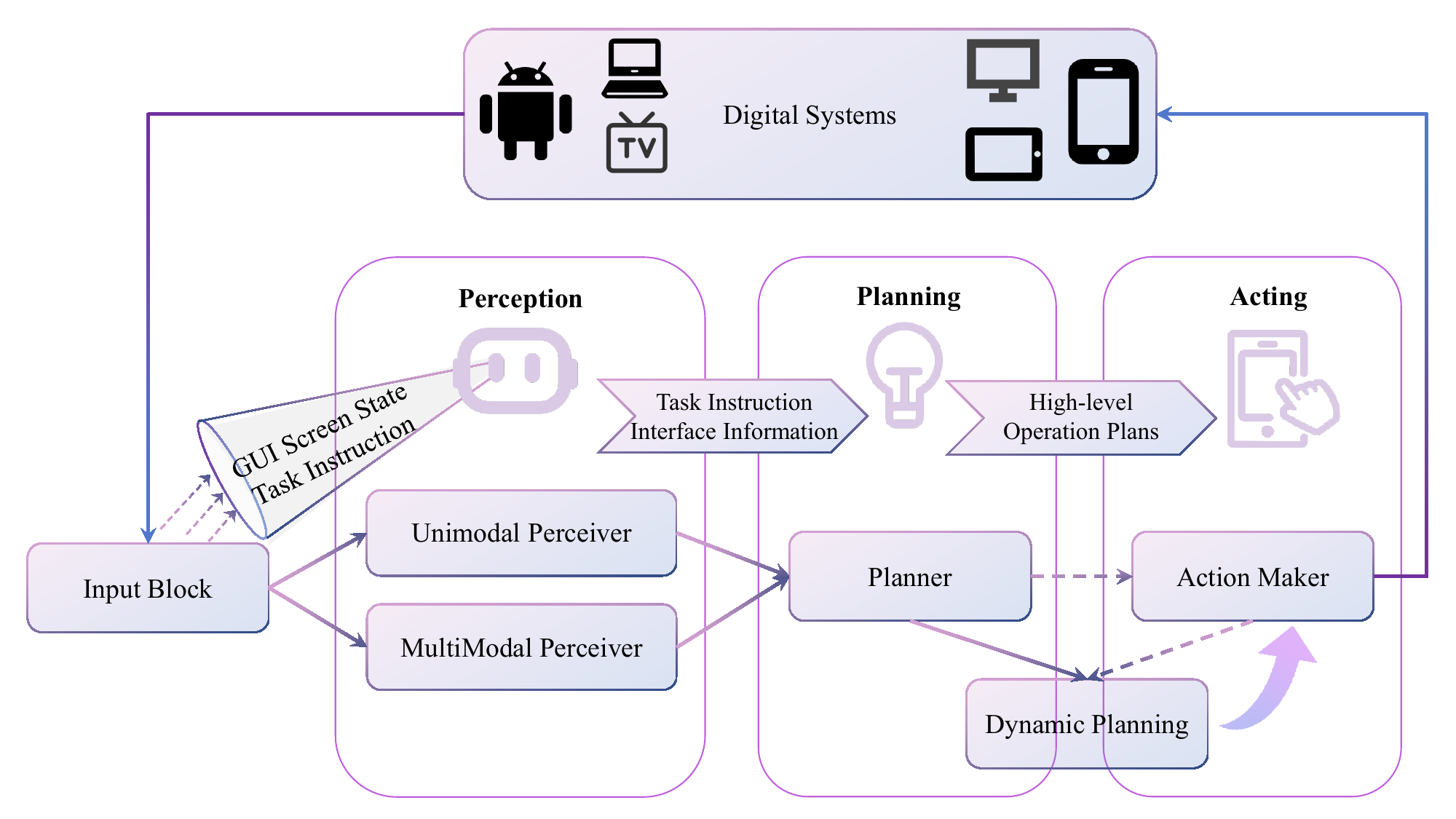}
    \caption{\textbf{Architecture of GUI agents powered by (Multi-modal) Large Language Models.} Comprising three interconnected modules: Perception, Planning, and Acting. The Perception module is responsible for perceiving and understanding the GUI state, the Planning module formulates high-level action strategies, and the Acting module executes operations, collectively enabling intelligent interaction with the GUI.}
    \label{fig:architecture}
    \vspace{0pt}
\end{figure*}

{\color{gray}\hrule}
\begin{center}
\section{(M)LLM-based GUI Agent Architectures}
\bigskip
\end{center}
{\color{gray}\hrule}
\begin{multicols}{2}
GUI agents are expected to autonomously operate digital devices operations to meet human task requirements. Typically, GUI agents receive a task instruction and the current screen state as input, and generate a sequence of planned actions to achieve the objective. As depicted in Figure ~\ref{fig:architecture}.The architecture generally consists of three key modules: Perception, Planning, and Acting.

\subsection{Perception}
The Perception module is responsible for perceiving and understanding the GUI by extracting semantic information from interactive elements like buttons, text boxes, and icons, laying the groundwork for subsequent planning and action execution. For unimodal language models, the GUI perception module mainly relies on accessible APIs provided by operating systems or applications, extracting the type and position of interface elements via the view hierarchy and feeding this symbolic information into the language model. These methods offer structured interface data suitable for model processing, though their effectiveness heavily depends on the developers’ proper implementation. However, when dealing with highly dynamic elements such as custom drawing canvases or game environments, the view hierarchy often fails to capture complete visual content, rendering the perception module ineffective. In contrast, for MLLMs, GUI agents can directly perceive the environment using visual information from screenshots. For example, OmniParser\cite{OmniParser} uses existing MLLMs (such as GPT-4V) to convert screen captures into structured representations of User Interface(UI) elements, extracting interface information directly from visual inputs and avoiding reliance on Accessibility APIs. As MLLMs continue to evolve, GUI perception has increasingly incorporated multimodal input sources for richer understanding. MP-GUI\cite{MP-GUI} introduces three specialized perceivers: a Textual Perceiver (TxP) for extracting text information, a Graphical Perceiver (GaP) for detecting graphical elements such as icons and buttons, and a Spatial Perceiver (SpP) for modeling spatial relationships between elements. By fusing the outputs of these perceivers, the model gains a more comprehensive understanding of the interface. Furthermore, CogAgent\cite{CogAgent} introduces a lightweight high-resolution cross-attention module that supplements the original low-resolution large image encoder with a high-resolution small image encoder, enhancing perception of complex GUI interfaces through cross-attention with the underlying Vision-Language Models(VLMs). Real-world GUI environments are dynamic, requiring perception modules to handle interface changes (e.g., loading animations, pop-ups) and noise (e.g., ads, notifications). UI-Hawk\cite{uihawk} highlights the use of historical screen context in GUI navigation, leveraging history-aware visual encoders and efficient resampling modules to enhance comprehension of changing interfaces. Additionally, the perception of screen visuals may raise privacy concerns due to the presence of sensitive information.

\subsection{Planning}
In modern (M)LLM-driven GUI agent architectures, the Planning module is responsible for translating perceived interface information and user task requirements into concrete action plans. With the development of VLMs, recent studies have explored the potential of using these models for planning and reasoning. RL4VLM\cite{zhai2024} introduced an approach that integrates reinforcement learning with chain-of-thought (CoT) reasoning in VLMs, where the model first outputs intermediate reasoning steps before the final action, and is fine-tuned using environment-based reward signals. This method employs CoT reasoning to guide the VLM in generating intermediate reasoning steps, leading to the final text-based action output. To further refine the planning process, Mobile-Agent-v2\cite{Mobile-Agent-v2} introduced an additional module within their multi-agent framework to generate more detailed execution plans. The system decouples the roles of the Planner, Decision Maker, and Reflector: the Planner outlines task progression to streamline navigation based on prior actions. The Reflector reviews execution outcomes after each step, providing feedback to dynamically adjust the plan. The recent dynamic planning method, D-PoT\cite{D-PoT}, breaks the traditional two-stage process of pre-defined planning followed by execution and feedback, integrating planning and feedback into a unified process: Following each action, the model captures the latest screen state and execution history, continuously revising the action sequence in real time until the task is successfully completed.
\subsection{Acting}
The GUI Acting module is responsible for converting high-level operation plans generated by the planning module into executable interactive actions on the interface, such as clicking, swiping, and text input [GUI-Odyssey\cite{GUI-Odyssey}; AppAgent\cite{AppAgent}. The design of this module directly affects the execution efficiency and task success rate of agents in real-world environments. UI-R1\cite{UI-R1} introduces a carefully designed unified reward mechanism and integrates multimodal inputs to guide the model in learning generalizable operational strategies for mobile interfaces. This method significantly improves performance in action type recognition and target localization, especially in identifying action types and locating UI elements. Subsequently, dynamic decision-making and history-aware mechanisms were incorporated into the Acting module to better handle long-horizon tasks and environmental changes. For instance, the D-PoT framework\cite{D-PoT} incorporates the concept of "Dynamic Planning of Thoughts" into the action generation process. After each action execution, the agent updates its remaining plan based on the latest interface feedback and execution history, enabling "plan-as-you-execute" behavior. This mechanism not only enhances adaptability in execution but also significantly improves the completion rate of complex tasks. To further improve generalization and cross-platform adaptability, GUI-R1\cite{GUI-R1} proposed the Group Relative Policy Optimization (GRPO) algorithm, which models shared strategies across tasks, enabling a single model to be applicable across platforms such as Windows, Linux, Android, and Web. Auto-GUI\cite{Auto-GPT} introduces a multi-modal chain-style action generation mechanism, combining visual and textual information to interact directly with the interface, thus avoiding the need for environment parsing or reliance on APIs. Overall, the Acting module is evolving from static mapping to an advanced module equipped with dynamic planning, history awareness, cross-platform adaptability, and end-to-end reasoning capabilities, with reinforcement learning, multimodal fusion, and module coordination becoming its core driving forces.

\begin{figure*}[thb]
    \centering
    \includegraphics[width=0.99\textwidth]{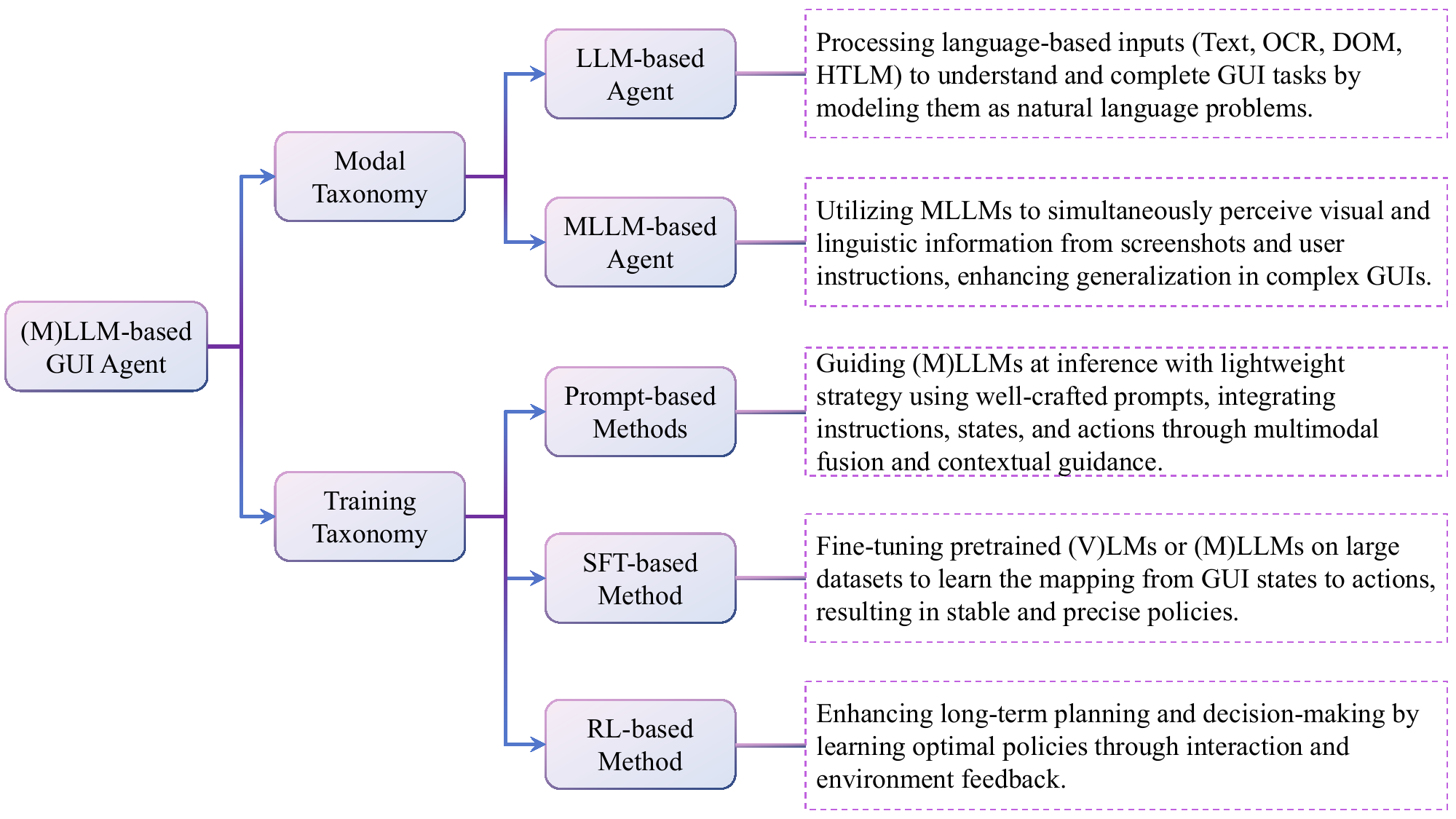}
    \caption{\textbf{Taxonomy of (M)LLM-based GUI Agents.} This figure presents a classification of GUI agents based on their underlying model modalities and training methodologies.}
    \vspace{10pt}
    \label{fig:Taxonomy}
\end{figure*}

\end{multicols}
{\color{gray}\hrule}
\begin{center}
\section{(M)LLM-based GUI Agent Taxonomy}
\bigskip
\end{center}
{\color{gray}\hrule}
\begin{multicols}{2}
With the development of MLLMs, the perception mechanisms of GUI agents have evolved from single-modality (e.g., text) to multimodal (e.g., a combination of vision and language). At the same time, training paradigms have diversified—from simple prompt-based guidance, to supervised fine-tuning, and further to RL-based policy optimization—providing strong technical support for GUI agents in diverse application scenarios. Figure~\ref{fig:Taxonomy} shows the taxonomy of GUI agents enhanced by foundation models. 

\subsection{Modal Taxonomy}
\subsubsection{LLM-based Agent}
Early GUI agents primarily relied on LLMs to perform tasks. These approaches were unimodal, as the agents could only process language-based inputs such as user task descriptions, screen text extracted via OCR, or structured DOM/HTML data to understand and complete tasks. This typically required GUI agents to first convert the screen content into text-based inputs. Therefore, during the perception stage, the screen had to be parsed and translated into a set of object descriptions. The core idea behind such methods is to model GUI tasks as problems of natural language understanding and generation. The LLM learns to select appropriate actions in different contexts through prompting, fine-tuning, or reinforcement learning. Socratic Models\cite{Socratic} proposed transforming structured environmental data (e.g., HTML trees of web pages) into natural language prompts, from which the LLM generates operation steps, thereby mapping "language to action". WebGPT\cite{WebGPT} integrated LLMs with search engine APIs to perform web-based question answering tasks. While LLM-based approaches demand relatively low data and deployment costs, they struggle to directly handle visual signals (e.g., button shapes, colors), have limited capacity to model interface structures, and thus suffer from poor generalization, making them difficult to apply to complex real-world graphical interfaces.
\subsubsection{MLLM-based Agent}
GUI interfaces are essentially multimodal interactive systems, consisting of visual components such as buttons, menus, and input boxes, which may not contain easily extractable structured textual information. As a result, traditional LLM-based approaches heavily rely on OCR and DOM structures, making it difficult to generalize to real-world apps or web applications. With the rapid development of MLLMs, GUI agents powered by MLLMs have increasingly become a focus of research in this field. Compared to traditional language-only methods, MLLMs can simultaneously perceive both visual and linguistic information, allowing them to complete tasks directly based on screen captures (or video frames) and textual instructions. This capability significantly enhances agents’ generalization and adaptability in complex GUI environments. For example, WebGUM\cite{WebGUM} introduced a multimodal task decomposition capability, where the model takes screen images and task instructions as input and performs staged perception, intent reasoning, and action generation, achieving notable results on MiniWoB++ and real-world web tasks. SeeAct\cite{SeeAct} introduces a vision-language agent capable of visual context awareness and action localization, using MLLMs like GPT-4V to reason over screen images and generate semantically relevant click or input actions. UI-TARS\cite{UI-TARS} proposed an agent architecture that combines native system operations with MLLM perceptual capabilities, emphasizing high-precision recognition and decision-making for complex GUI components in the Android system. Using screen snapshots and task descriptions, the model applies vision-language reasoning for holistic perception and operational planning.
\subsection{Training Taxonomy}
\subsubsection{Prompt-based Methods}
Prompt-based methods offer a lightweight strategy for building GUI agents by harnessing the pretrained strengths of (M)LLMs. Through well-crafted prompts, these models are guided to perform tasks at inference time without parameter tuning or high compute resources. The core of prompt-based methods lies in structured prompt engineering, which integrates user instructions, environment states, and action specifications into model-understandable inputs through multimodal fusion and contextual guidance. Zeng et al.[2022] proposed the Socratic Models framework\cite{Socratic}, which uses multi-round prompts to convert visual or structured data (e.g., HTML trees or OCR-extracted text) into natural language, enabling LLMs to output actions and complete zero-shot multimodal tasks such as video QA and robot planning. Subsequently, researchers extended prompt-based approaches to real-world web automation. WebAgent\cite{WebAgent} incorporates HTML summaries and Python subprogram generation in its prompts, breaking long HTML documents into task-relevant segments and generating executable code for end-to-end web automation. With the rapid advancement of MLLMs, prompt-based approaches have further evolved to take screenshots directly as input. UFO2\cite{UFO2} integrates VLMs and LLMs to construct multimodal inputs (screenshots + Accessibility API control attributes), generating cross-platform action sequences through hierarchical prompts to support automation across Windows, web, and mobile platforms.
\subsubsection{SFT-based Method}
Supervised Fine-Tuning (SFT) is a widely adopted training paradigm for GUI agents. Unlike zero-shot or few-shot learning, this approach fine-tunes pretrained (V)LMs or MLLMs on large annotated datasets, enabling models to learn the mapping from GUI states to concrete actions, resulting in more stable and precise behavioral policies tailored to specific GUI tasks. SFT offers advantages such as a stable training process and effective specialization for target tasks. Auto-GUI\cite{Auto-GPT} applies supervised fine-tuning on high-quality GUI interaction datasets (including screenshots, task descriptions, action histories, and annotated next-step actions), using GUI states and language tasks as input to predict the next action—such as action type, click coordinates, and text input. This method achieved remarkable performance on several GUI agent benchmarks, such as AITW and Mind2Web. Some efforts focus on building general-purpose GUI understanding models first, and then applying SFT on downstream tasks. For example, GUICourse\cite{GUICourse} pretrains general VLMs (like Qwen-VL and MiniCPM-V) on the GUIEnv dataset to enhance OCR and element localization; Falcon-UI\cite{Falcon-UI} is pretrained on the Insight-UI dataset to enhance the model's understanding of GUI environments. This model are then fine-tuned on Android and Web GUI datasets (e.g., AITW, AITZ, Android Control, and Mind2Web), achieving performance comparable to larger models like Qwen2VL; InfiGUIAgent\cite{InfiGUIAgent} proposes a two-stage supervised fine-tuning framework. The first stage boosts GUI perception and localization, while the second uses synthetic data to teach reasoning and self-correction in complex multi-step scenarios. Additionally, many works focus on dataset construction and task diversity. TongUI\cite{TongUI} constructs the GUI-Net dataset, containing 143K interaction records, by automatically crawling GUI operation traces from online tutorials. In conclusion, SFT-based methods continuously improve model robustness and task success rates in real GUI environments by expanding data scale and diversity and adopting staged training processes.
\subsubsection{RL-based Method}
RL is primarily used in GUI agent training to enhance long-term planning and decision-making capabilities in complex task scenarios. Recent research highlights RL's crucial role in enabling agents to learn optimal policies via interaction, particularly in handling intricate, multi-step tasks to boost decision-making and task success. Unlike SFT, which depends on human-labeled expert data, RL explores and learns policies through trial with environment feedback. This makes it well-suited for real GUI scenarios with long task sequences, ambiguous goals, and sparse feedback. 

Recently, many works have integrated RL into MLLMs. RL4VLM\cite{zhai2024} applies Proximal Policy Optimization (PPO) to VLMs, generating CoT reasoning at each step, converting the output into actions, receiving environment rewards and fine-tuning the VLMs accordingly. To balance data efficiency and cross-platform generalization, GUI-R1\cite{GUI-R1} employs RL to enhance the GUI operation capabilities of VLMs in complex real-world scenarios. It incorporates DeepSeek-R1-style "regularized rewards" into GUI action prediction, using a unified action space and the Group Relative Policy Optimization (GRPO) algorithm to train across platforms such as Windows, Linux, MacOS, Android, and Web. For long-horizon tasks spanning multiple apps or APIs, short-term RL often suffers from local optima. Chen et al. [2025] propose LOOP\cite{LOOP}, a memory-efficient variant of PPO that trains IDAs (Intelligent Digital Agents) directly in target environments. It avoids value networks and requires only a single LLM instance in memory to perform end-to-end RL training within the AppWorld environment. Due to the high cost and brittleness of real environment interaction, environment-free RL approaches have emerged. Zheng et al. (2025) propose VEM (Value Environment Model)\cite{VEM}, where a value function is trained on offline GUI interaction data to estimate long-term returns for arbitrary state-action pairs, serving as “virtual rewards” to guide policy optimization. In addition, asynchronous interaction approaches have been incorporated into RL training\cite{distrl}\cite{Wang_2024}. Trajectories generated by GUI agents in simulated environments are stored in a replay buffer, serving as data for subsequent online model training.
\end{multicols}
\clearpage

{\color{gray}\hrule}
\begin{center}
\section{Challenges}
\bigskip
\end{center}
{\color{gray}\hrule}
\vspace{0.5cm}
\begin{multicols}{2}
Despite the rapid progress in (M)LLM-based GUI agents, several challenges remain:

\noindent \textbf{Perception under Dynamic and Noisy Interfaces:}\quad 
Real-world GUIs are often dynamic, featuring pop-ups, advertisements, or frequent layout changes. Although modern GUI Perception module integrate multimodal inputs and history-aware modeling, accurately perceiving and adapting to these unpredictable variations remains a major difficulty.

\noindent \textbf{Long-Horizon Planning and Execution:}\quad 
GUI tasks typically require multiple sequential operations. Ensuring consistency across multi-step plans, especially when intermediate states deviate from expectations, challenges current Planning modules. Dynamic replanning strategies\cite{D-PoT} have made progress, but long-horizon scalable and reliable reasoning remains open.

\noindent \textbf{Data Efficiency and Generalization:}\quad 
SFT-based approaches heavily rely on large annotated datasets. Despite advances like GUI-R1\cite{GUI-R1} using small datasets, achieving strong generalization across different platforms (Windows, Android, Web) with minimal supervision is still an ongoing challenge.

\noindent \textbf{Evaluation and Benchmarking:}\quad 
Current benchmarks like Mind2Web and AITW primarily focus on success rates. However, finer-grained evaluations (e.g., plan optimality, robustness under interface drift) and standardized human-in-the-loop assessments are urgently needed to fully measure agent capabilities.
\end{multicols}
{\color{gray}\hrule}
\begin{center}
\section{Conclusions}
\bigskip
\end{center}
{\color{gray}\hrule}
\vspace{0.5cm}
This paper presents a structured summary of recent developments in GUI agents enhanced by foundation models and RL techniques. We first introduced the task formulation, key execution environments, and standard evaluation metrics. Then, we reviewed the modular architecture of GUI agents — covering Perception, Planning, and Acting — along with representative advances in each component. We also discussed three major training paradigms: Prompt-based methods for lightweight deployment, SFT-based methods for domain-specific adaptation, and RL-based methods for dynamic policy learning.

Through this review, we observe a clear trend: GUI agents are evolving from static rule-based systems toward adaptive, perception-driven, and reasoning-capable agents, powered by MLLMs and optimized via RL. Despite impressive gains, challenges in perception robustness, long-horizon reasoning, data efficiency, and evaluation remain to be addressed.

Looking forward, integrating better semantic grounding, continual learning, human feedback, and asynchronous interaction will be crucial for building GUI agents that are truly capable of autonomously navigating complex, dynamic digital environments.
\bibliographystyle{plain}
\bibliography{references}
\end{CJK}
\end{document}